# A global product of fine-scale urban building height based on spaceborne lidar

Xiao Ma[1], Guang Zheng[2], Chi Xu[1], L. Monika Moskal[3], Peng Gong[4], Qinghua Guo[5], Huabing Huang[6], Xuecao Li[7], Yong Pang[8], Cheng Wang[9], Huan Xie[10], Bailang Yu[11], Bo Zhao[12], Yuyu Zhou[4]

[1] School of Life Sciences, Nanjing University, Nanjing 210023, China

[2] International Institute for Earth System Sciences, Nanjing University, Nanjing 210023, China

[3] School of Environment and Forest Sciences, University of Washington, Box 352100, Seattle, Washington 98195, USA   lmmoskal@uw.edu

[4] Urban Systems Institute, Departments of Geography and Earth Sciences, University of Hong Kong, Hong Kong 999077, China   penggong@hku.hk and   yuyuzhou@hku.hk

[5] Institute of Remote Sensing and Geographical Information Systems, School of Earth and Space Sciences, Peking University, Beijing 100871, China   guo.qinghua@pku.edu.cn

[6] School of Geospatial Engineering and Science, Sun Yat-Sen University, and Southern Marine Science and Engineering Guangdong Laboratory (Zhuhai), Zhuhai 519082, China   huanghb55@mail.sysu.edu.cn

[7] College of Land Science and Technology, China Agricultural University, Beijing 100083, China   xuecaoli@cau.edu.cn

[8] Institute of Forest Resource Information Techniques, Chinese Academy of Forestry, Beijing 100091, China   pangy@ifrit.ac.cn

[9]Key Laboratory of Digital Earth Science, Aerospace Information Research Institute, Chinese Academy of Sciences, Beijing 100094, China, wangcheng@aircas.ac.cn

[10]College of Surveying and Geo-Informatics, and Shanghai Key Laboratory of Space Mapping and Remote Sensing for Planetary Exploration, Tongji University, Shanghai 200092, China   huanxie@tongji.edu.cn

[11]School of Geographic Sciences, East China Normal University, Shanghai 200241, China blyu@geo.ecnu.edu.cn

[12]Department of Geography, University of Washington, Smith Hall 408, Box 353550, Seattle, WA 98195, USA   zhaobo@uw.edu

**Authors for correspondence**:

Guang Zheng (zhengguang@nju.edu.cn) and Chi Xu (xuchi@nju.edu.cn)

**Abstract:**

Characterizing urban environments with broad coverages and high precision is more important than ever for achieving the UN's Sustainable Development Goals (SDGs) as half of the world's populations are living in cities. Urban building height as a fundamental 3D urban structural feature has far-reaching applications. However, so far, producing readily available datasets of recent urban building heights with fine spatial resolutions and global coverages remains a challenging task. Here, we provide an up-to-date global product of urban building heights based on a fine grid size of 150 m around 2020 by combining the spaceborne lidar instrument of GEDI and multi-sourced data including remotely sensed images (i.e., Landsat-8, Sentinel-2, and Sentinel-1) and topographic data. Our results revealed that the estimated method of building height samples based on the GEDI data was effective with 0.78 of Pearson's *r* and 3.67 m of RMSE in comparison to the reference data. The mapping product also demonstrated good performance as indicated by its strong correlation with the reference data (i.e., Pearson's *r* = 0.71, RMSE = 4.60 m). Compared with the currently existing products, our global urban building height map holds the ability to provide a higher spatial resolution (i.e., 150 m) with a great level of inherent details about the spatial heterogeneity and flexibility of updating using the GEDI samples as inputs. This work will boost future urban studies across many fields including climate, environmental, ecological, and social sciences.

# 1. Introduction

Cities are currently home to ~45 billion people, or ~57% of the world's total population, and these numbers are projected to continue growing in the coming decades (United Nations. 2022). Achieving sustainable, resilient cities as the major habitat of our species is therefore a crucial goal for sustaining the existence of human civilizations in the face of increasingly severe climate change and other impacts (Pedersen Zari et al. 2022). The rapid development of cloud computing services, multiple remote sensing observations, and artificial intelligence in recent years has boosted our ability to characterize, understand, and monitor urban systems with big data (Amani et al. 2020; Patino and Duque 2013). However, high-precision urban data are mostly available for the most developed megacities, and major gaps remain for many other cities across the world, especially in Global South countries, creating a key inequality that hampers the UN's Sustainable Development Goals (Herfort et al. 2023; Wang et al. 2023).

Three-dimensional urban structure, such as building height, is fundamental information about urban environments and has far-reaching applications for scientific research and management of urban systems (Dimoudi and Nikolopoulou 2003; Li et al. 2020b; Perini and Magliocco 2014). With the increasing availabilities of remotely sensed data and topographic data, growing studies try to map building heights at the regional or global scale based on direct and indirect methods. The direct method generates the global building height maps from the normalized digital surface model (nDSM), which is mainly derived from stereo remotely sensed images (e.g., GaoFen-7) or digital elevation model (DEM) products (e.g., ALOS DSM). Esch et al. (2022) first produced the world settlement footprint 3D map at a 90-m grid size using the TanDEM-X product. He et al. (2023) used the ALOS DSM product to generate the 30-m global building height. Due to the limited DEM products, most previous studies chose to produce the building height map by calibrating the regression relationship between the building height samples and various features derived from the multi-source spatial data. For instance, Li et al. (2022) built random forest models to map global built-up 3D patterns at the 1-km resolution around 2015 including the building height parameter. Based on mapping global built-up height using the Sentinel-1 data, Zhou et al. (2022) revealed the large inequality in infrastructure availability in the Global South. A notable advance is the recent production of a global building height distribution map from the Global Human Settlement Layer-Built-Height (GHSL-H) dataset at a 100-m grid size for the year 2018 (Pesaresi and Politis 2023), which mainly estimated the building height using the DEM-derived features (i.e., ALOS DSM and SRTM DEM) and updated the last predictions from the Sentinel-2 data.

However, existing global products are subject to relatively coarse spatial resolutions (e.g., 500 m or 1 km) and time lags. Especially, the nDSM-derived global building height products show limitations in the analysis of the current urban environment, which mainly used the DEM products about ten years ago (Frantz et al. 2021). Although the new Chinese multi-view satellite of GaoFen-7 performed greater potential in estimating the high-resolution building height map, the GaoFen-7 images haven't full coverage in the world (Chen et al. 2023). Moreover, in the indirect approach, researchers paid more attention to exploring effectively explanatory variables and developing various regression algorithms, such as adding the nighttime light (NTL) data and employing the deep learning method (Cai et al. 2023; Cao and Huang 2021; Wu et al. 2023). Nevertheless, obtaining global continuous distribution of building height samples is also challenging as they were produced from very different sources, such as digital elevation model (DEM) data, open GIS data, and lidar (light detection and ranging) data, with contrasting levels of accuracy and uncertainty (Geiß et al. 2020; Li et al. 2022).

Spaceborne lidar is so far the most promising tool for sampling building heights with close-to global coverages. Particularly, the recently launched Global Ecosystem Dynamics Investigation (GEDI) instrument has shown great power in retrieving 3D structures on the ground using footprint-based measurements - while GEDI was originally designed to measure forest vertical structure with high sampling density concerning carbon storage and deforestation (Dubayah et al. 2022; Kellner et al. 2023; Potapov et al. 2021). It is undoubtedly a useful tool for mapping urban building heights at the 150-m spatial resolution (Ma et al. 2023). Compared with the complex noise filtering and photon classification process of the ICEsat-2 data, (Dandabathula et al. 2021; Lao et al. 2021; Zhao et al. 2023), the GEDI-based relative height metrics are more straightforward and simple in building height estimation.

Here we provide an up-to-date global product of urban building height at the 150-m spatial resolution around 2020 through extrapolating GEDI-based building height samples using wall-to-wall time-series features of spectral, backscattering coefficients about buildings and topography information across the world. The newly proposed method stemmed from our previously developed GEDI-based regional building height mapping method that has been successfully applied and tested in the Yangtze Delta region in China (Ma et al. 2023). We further improved the previous method at the GEDI footprint level by developing random forest (RF) models to correlate between GEDI-derived building height samples and the variables from spatially continuous remotely sensed imageries for 68 subregions at a global level, respectively. We then employed these models to obtain a global map of spatially continuous urban building heights. To conduct the validation and comparisons, we collected the reference data from North America, Europe, China, Brazil, and New Zeeland and the recently released 100-m and 1-km global building height maps produced by Pesaresi and Politis (2023) and Li et al. (2022), respectively. The global product of urban building heights is available by following the website (https://nju-eco-lidar.projects.earthengine.app/view/gbh2020) powered by the Google Earth Engine (GEE).

## 2. Materials and methods

**2.1 Data and Processing**

We collected five different categories of information to build up the statistical models between GEDI-based building heights as the dependent variable and various remotely sensed features about buildings including time-series imageries (i.e., Landsat-8, Sentinel-2, and Sentinel-1) and topographic information and assess the final map based on the reference building height and auxiliary data (Table 1). We generated the building height samples from the GEDI L2A Version 2 data. Using the Landsat-8 and Sentinel-2 optical imageries ranging from January 2020 to December 2021, we computed the multitemporal statistical and textural features based on the spectral indices. To alleviate the effects of cloud and cloud shadows, we also produced similar multitemporal statistical and textural features based on the raw bands of time-series Sentinel-1 imageries for the whole year of 2020. The built-up areas of final mapping were generated based on the Global Urban Boundary 2018 (GUB2018) (Li et al. 2020a). In addition, we also collected the reference data of building heights in the forms of both raster and vector for validation purpose. All remotely sensed data were acquired and processed based on the Google Earth Engine (GEE) (Gorelick et al. 2017).

Table 1. Multi-sourced datasets applied in the global urban building height mapping

| Category | Data Source | Features/Description |
|---|---|---|
| Spacebrone lidar | GEDI L2A | Generate building height samples |
| Optical images (2020.01~2021.12) | Landsat-8 Sentinel-2 | Spectral indices (NDVI, NDBI, mNDWI, DVI64, RVI64) <br> Multitemporal Statistic Features (mean, variance, percentiles (i.e., p10/p25/p90)) *spectral indices* <br> Texture Features (mean, variance, contrast, and dissimilarity) *multitemporal statistic features* |
| Radar images (2020.01~2020.12) | Sentinel-1 | Radar indices (VV, VH) <br> Multitemporal Statistic Features (mean, variance, percentiles (i.e.,p0/p10/p25/p90/p100)) *radar indices* <br> Texture Features (mean, variance, contrast, and dissimilarity) *multitemporal statistic features* |
| Terrain data | ASTER GDEM | Elevation, Slope, Aspect |
| Reference data | Normalized DSM <br> Building footprint with height information | Validation |
| Auxiliary data | World Settlement Footprint 2019 (WSF 2019) <br> Building Footprint <br> Global Urban Boundary 2018 (GUB 2018) | Mask out non-building information <br> Filter GEDI footprint data <br> Mapping and Analysis |

**2.1.1 GEDI**

The global GEDI footprints covering the world's human settlement regions were first derived from the mouth raster GEDI L2A product ranging from 2019 to 2021 by synthesizing the monthly orbits of the raw GEDI L2A product (Dubayah et al. 2021) with geolocation, root-level relative height, and quality flags information within the GEE platform. With the cross-track pointing ability of the GEDI system (Dubayah et al. 2020), we assume that the laser footprint has a lower probability of covering the same area each month and convert the mouth raster GEDI L2A layer into a vector point layer as GEDI footprints to represent the samples of vertical structures within global coverages. We filtered out the low-quality (i.e., quality flag = 0 and sensitivity < 0.9) or potentially deteriorated (i.e., degrade flag > 0) footprints. We also masked out all GEDI footprints falling outside built-up areas using the World Settlement Footprint data 2019 (Marconcini 2021). We projected the rest high-quality footprints into the World Mollweide projection to reduce distance distortion in high-latitude areas for the further process (Fig. 1a1). Following the similar procedures of our previous method (Ma et al. 2023), we produced a high-quality GEDI footprints vector dataset with the 85th (RH85) and 95th (RH95) percentiles of relative height metrics in each footprint for all built-up areas in the world.

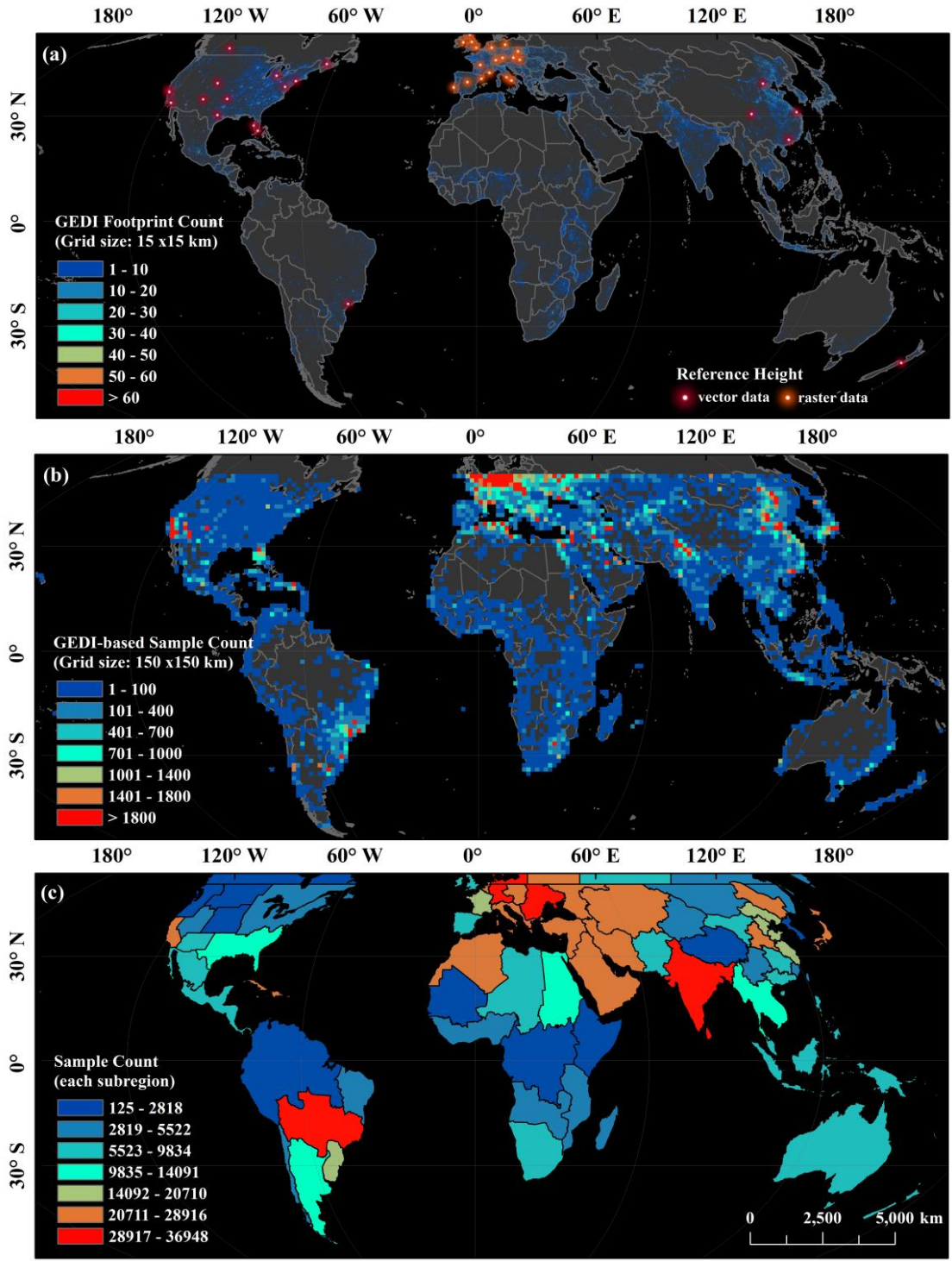

**Fig. 1.** The global urban region covered with the quality-sifted GEDI footprint data and the reference building height data (a); the density distribution of GEDI-based building height samples with a 150-km grid (b); and the number of building height samples of each subregion model (c).

**2.1.2 Time series datasets**

We collected optical remotely sensed imageries of Landsat-8 and Sentinel-2 that are atmospherically corrected land surface reflectance data using the Land Surface Reflectance Code (LaSRC) and the Sen2Cor processer. For each image, the clouds and their shadow pixels were masked out based on the quality and S2-cloud-probability bands of the Landsat-8 dataset and Sentinel-2, respectively. We only selected the Landsat-8 and Sentinel-2 imageries with less than 30 % and 20 % cloud coverage to compute the spectral indices including NDBI (Zha et al. 2003), NDVI (Rouse 1973), mNDWI (Xu 2006), DVI64, and RVI64. In this study, we used five spectral features and two dual-polarization bands (i.e., VV and VH) of Sentinel-1 imageries for generating independent features in the next processing.

**2.1.3 DEM and auxiliary data**

The ASTER GDEM V3 was also collected in the current study to represent the global topographic variations covering most urban areas with latitudes ranging from 83°N to 83°S (Abrams et al. 2020; Abrams et al. 2022). We masked the non-building areas to protrude independent features related to building heights using the World Footprint Settlement 2019 product (Marconcini 2021). Compared with the existing urban boundary maps produced using remotely sensed data with coarse resolution, the GAIA-derived global urban boundary datasets (GUB2018) hold the promising ability to exhibit a high level of detail about the city's extent and boundary (Li et al. 2020a). It was applied to determine the mapping areas in this study. For selecting GEDI footprints with a high probability of capturing the relative heights of buildings from waveforms and generating building height samples from the sifted GEDI data, we collected the essential building footprint data around the world from Microsoft Bing Map (https://github.com/microsoft/GlobalMLBuildingFootprints), Tiandi Street Map (https://map.tianditu.gov.cn/), building footprints of partial Chinese cities (Zhang et al. 2022) and Open Street Map (OSM) (https://download.geofabrik.de).

**2.1.4 Reference data**

To validate GEDI-based building height samples and our global building height map, we collected the reference data for 39 cities in both vector (20 cities) and raster (19 cities) forms across North America, Europe, China, Brazil, and New Zealand. For North America, the open reference data collected from local governmental portals (http://hub.arcgis.com) consisted of airborne lidar data for Washington D.C. and 13 building footprint files with height recording for cities ranging from metropolis (e.g., New York, Miami, and Los Angeles et al.) to small towns in the rural areas. The European building heights data collected from Copernicus Global Land Service (2013) were produced by the 10-m normalized DSM derived from the IRS-P5 stereo images. The reference data of 4 cities (i.e., Beijing, Shanghai, Chengdu, and Guangzhou ) in China and Sao Paulo (Brazil) were recorded by the building footprint data with the floor number (https://www.amap.com) and height tag (www.openstreetmap.org), respectively. We obtained the reference data of Wellington City by calculating the difference between DSM and the digital terrain model (DTM) that were acquired from the open topographic website (https://www.opentopography.org/). The vector files of building footprints with height information were employed for assessing the GEDI-derived building heights and global product. The 10 m grid size raster data were used to assess the final global building map due to their limited ability to identify the individual buildings compared with the vector building footprint layer associated with height information.

**2.2 GEDI-based building height samples**

We created global building height samples by aggregating filtered high-quality GEDI footprints into a 150-m grid size. We estimated building heights using a composite relative height index (RH) at footprint level by: (a) setting the GEDI-based RH95 metrics as primary building height; (b) replacing the building height with the RH85 metrics for some areas with high vegetation coverage; (c) and saving all footprints with the values of RH were greater than 2.5 m. By doing so, we computed the mean values of the RH index of all high-quality footprints (count $\geqslant$ 3) located in each grid with a 150-m size as the actual mean structure height and used them as the dependent variable to calibrate building height models. The uneven spatial distribution patterns of GEDI-based building height sample numbers could be observed ([Fig. 1a2](#)) ranging from less than 100 to more than 1,800 in each 150-km grid around the world. More samples are located in the regions of Europe, East Asia, North Africa, and southwestern North America based on the different building footprint densities, levels of urbanization, and orbit track design of GEDI aboard platform (i.e., the International Space Station).

**2.3 RS-based explanatory features**

Five spectral feature variables (i.e., NDBI, NDVI, mNDWI, RVI64, and DVI64) and two radar bands (i.e., VV and VH) were used to compute the multitemporal statistical features (Table 1) to reduce the effects of inconsistent temporal observations by acquiring the phenological signals ([Frantz et al. 2021; Potapov et al. 2021](#)). The ability of effectively capturing the surrounding building information of texture features could obtain complete shadow features as much as possible from a 10-m or 30-m pixel ([Wu et al. 2023](#)). Therefore, we also computed four texture features (i.e., mean, variance, contrast, and dissimilarity) using the Grey Level Co-occurrence Matrixes ([Haralick et al. 1973](#)) based on the multitemporal statistic features. Following Ma et al. (2023), the window size of texture features were set as $2 \times 2$, $6 \times 6$, and $5 \times 5$ for Landsat-8, Sentinel-2, and Sentinel-1, respectively. Topographic variables include elevation, aspect, and slope. In total, 323 explanatory variables were involved in extrapolating discrete samples to the spatially continuous map.

**2.4 RF-based Global Urban Building Height Estimation**

In this study, we used the subregion modeling method to estimate building heights by combining GEDI-based building height samples, RS-derived features, and topographic features ([Fig. 2](#)). We divided each modeling area according into administrative regions and climate zones ([Beck et al. 2018](#)) to ensure that each mapping area had a similar context. The number of samples per subregion is shown in [Fig. 1c](#). Since GEDI does not cover high latitude areas above 51.6° N, we trained 5 models of the northern region based on the samples within 600 km nearby. Random Forest modeling was used to establish the regression relationship between the GEDI-based sample and independent features. All modeling processes were completed using the

Rstudio based on the *randomForest* package (Breiman et al. 2018). We saved all the features with the number 323 for each subregion modeling to keep the consistency of interpretation.

**2.5 Assessment Accuracy**

Indicators of Pearson's correlation coefficient (*r*) and root mean square error (RMSE) were used for accuracy assessment. First, we used the reference data consisting of building footprint vector files to evaluate the effectiveness of GEDI-based building height estimation at the footprint and sample levels. Then, we converted all the reference data to the Mollweide projection using the mapping product and generated the actual mean building height using the area-weighted and downscaling method respectively based on the vector and raster reference data to evaluate the building height map.

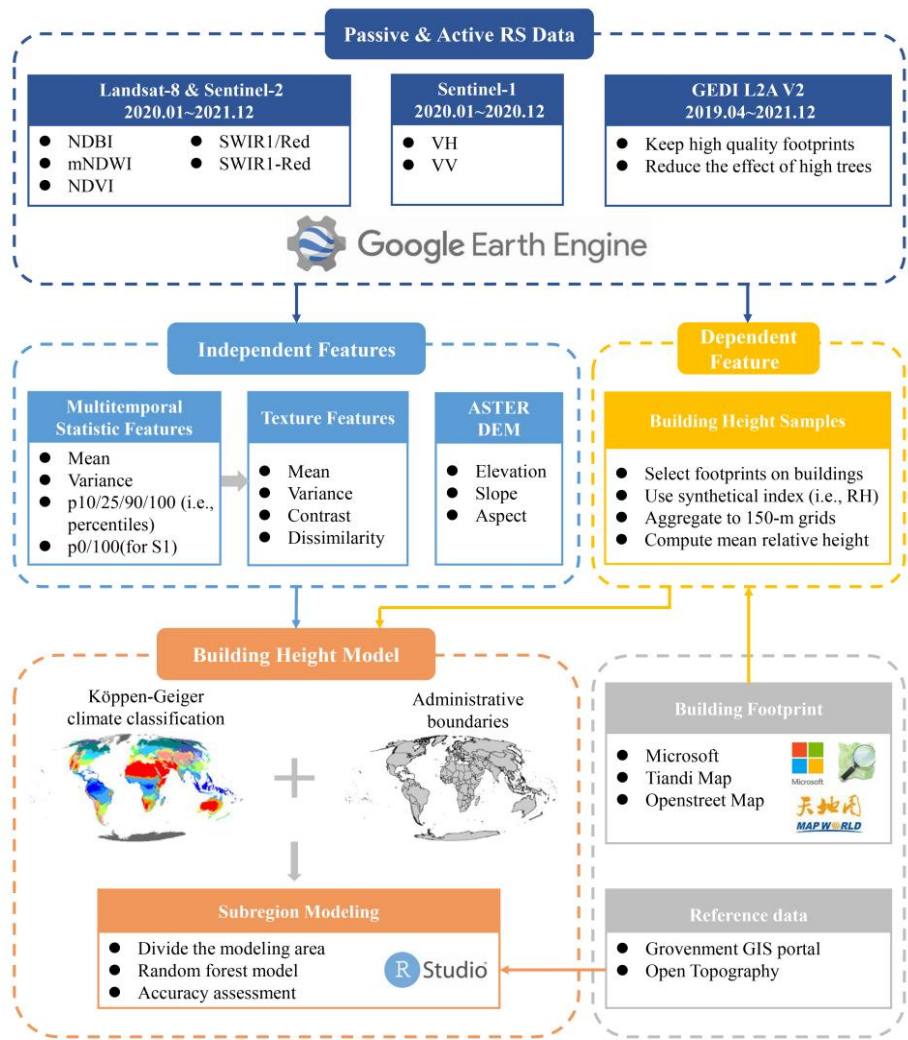

**Fig. 2.** The flowchart of global urban building height mapping.

## 3. Results and Discussion

**3.1 Validation of GEDI-based Building Height Estimation**

Our results showed that the RH index had good agreement with the reference data (Pearson's $r$ = 0.70, RMSE = 5.51 m). For the GEDI-derived building height samples, the mean values of the RH index performed a higher correlation relationship (i.e., Pearson's $r$ = 0.78) and lower error (i.e., RMSE = 3.67 m). The flattened ground around buildings, obvious three-dimensional structures, and mostly unified architectural forms in a block may explain the effectiveness of building height estimation based on GEDI data (Ma et al. 2023). However, the RH index still tended to overestimate the height of low-rise buildings, and GEDI-based samples may underestimate the actual mean building height in some areas (Figs. 3a&b). Large residuals may be caused by some issues with the GEDI V2 product like the geolocation error and the low footprint density (Ma et al. 2023).

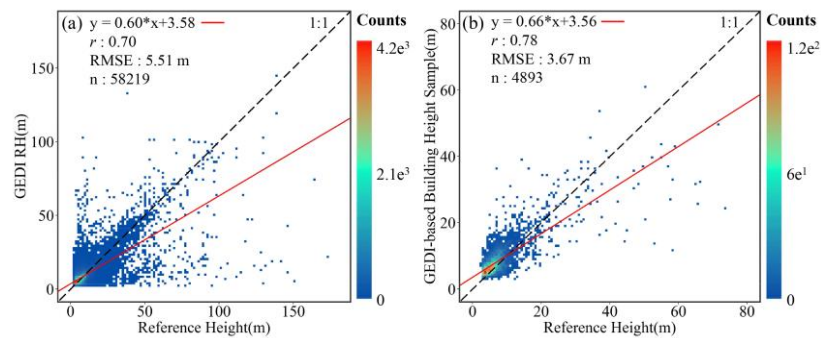

**Fig. 3.** Accuracy assessment of building height derived from GEDI data at the footprint level and the 150-m scale (a and b).

**3.2 Global and Regional Building Height Overviews**

The overview of medium-height cities (i.e., area ⩾ 10 km$^2$) of our product is shown in Fig. 4a through the 3D maps rendering at the 1.5-km grid size. The 3D views revealed that high-rise buildings were mainly found in East Asia (Fig. 4b3). The 2D/3D maps of cities with the 150-m scale could provide clear details of building height distributions (Fig. 5). For instance, the building height patterns of metropolitan cities usually exhibited a 3D form of a single peak with a gradual decrease from the center to the surrounding area in Europe, America, and

Australia (Figs. 5c1, c2, d1, d2, e1, e2). However, some cities showed clearly different 3D forms, such as Lagos with generally low buildings (Figs. 5a2-I&a2-II).

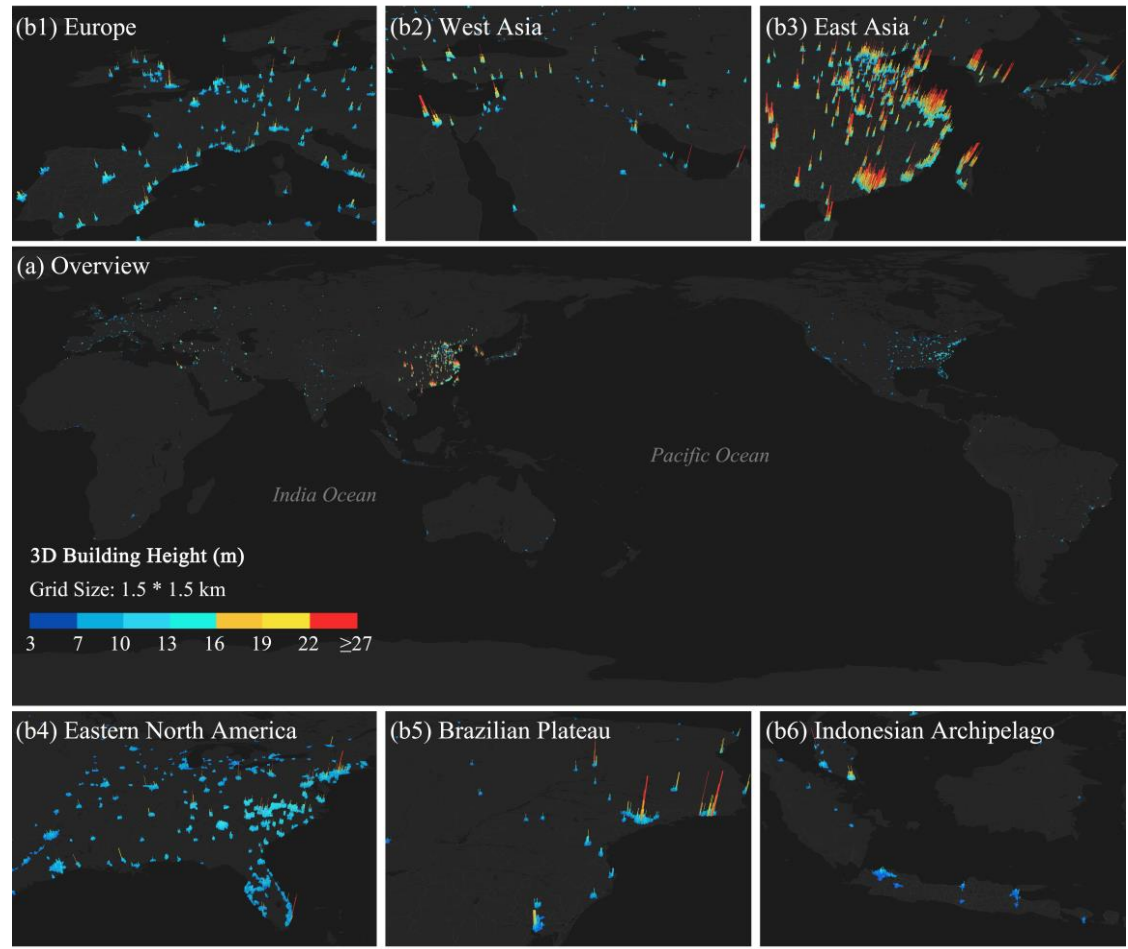

**Fig. 4.** 3D global overview of this urban building height map at the 1.5-km scale (a); close-up 3D views for six regions (b1-b6).

Furthermore, we also paid close attention to the close-up scenes with four-by-eight grids to validate the consistency of our product and the actual building height distribution. There were obvious long shadows of tall buildings in the corresponding areas of red pixels by referring to the very high resolution (VHR) Bing images (https://learn.microsoft.com/en-us/bingmaps/rest-services/directly-accessing-the-bing-maps-tiles). The actual region covered with blue pixels represented an opposite situation, which gathered a lot of low buildings with fewer shadows. Therefore, our map could correctly perform the spatial distributions of building heights for the global cities.

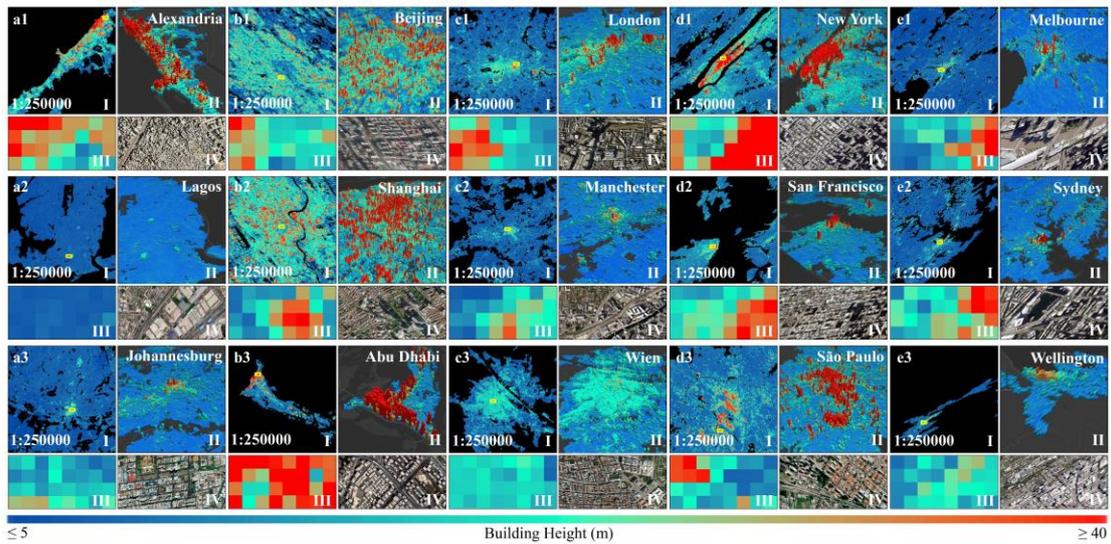

**Fig. 5.** 2D (a-e-I) and 3D (a-e-II) views of building heights of 15 cities at the 150-m scale; and comparisons between close-up 2D views (a-e-III) and the high spatial resolution Bing images (a-e-IV).

**3.2 Validation of product**

**3.2.1 Comparison to the reference data**

Based on the reference data, the evaluation of the building height map showed Pearson's *r* and RMSE values of 0.70 and 4.49 m, respectively (Fig. 6). The *r* and RMSE values were 0.69 and 3.91 m for the north region without the direct GEDI calibration. Figs. 3c&d reveal that the estimated heights could remain consistent with the actual data for building heights less than 20 m. However, the great uncertainty of the product could be observed for the height estimation of tall buildings (> 30 m). The saturation effect was also found in the related previous studies (Cao and Huang 2021; Frantz et al. 2021; Wu et al. 2023), which might be mainly explained by shortcomings of samples and independent features. As an aspect, the uncertainty and less high values of GEDI-based samples might impact the accurate calibration and the estimation ability per model. As the other aspect, unequally available satellite observations, limited explanatory ability, and masking loss effective pixels of optical or radar features may cause the underestimation for the dense areas of high-rise buildings (Frantz et al. 2021; Ma et al. 2023; Potapov et al. 2021).

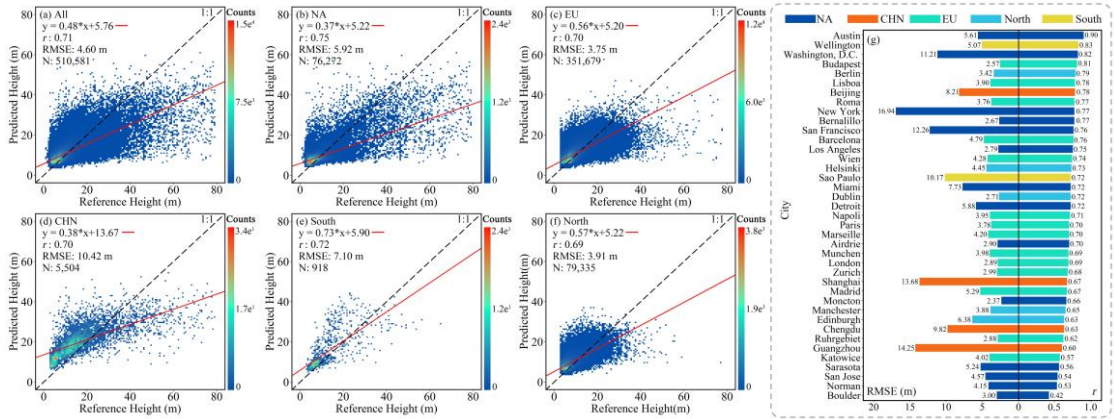

**Fig. 6.** Validation of global urban building height map and north map without GEDI-based samples using the reference building height data (a-f); the city-level validation of product (g).

**3.2.2 Comparison to the Existing Products**

We compared our map with two recently released global building height maps (i.e., GHSL-H and Li et al. (2022)), respectively. To keep the similar estimated results, the average net building height product of GHSL-H was selected for further comparison (Pesaresi and Politis 2023). Li et al. (2022) produced a global 3D map of built-up (i.e., Building height) at the 1-km$^2$ grid size for the nominal year 2015 by establishing the regression relationship between the reference data and features derived from multi-sourced spatial data.

We firstly resampled GHSL-H to 150-m resolution for comparing this product with our building height map in cities larger than 100 km$^2$ (Fig.7). In general, the correlation between the two products was lower with the *r* values from 0.9 to 0.54 at the pixel scale of each continent (Figs. 7a1-a6). In particular, the RMSE of two products of Asia cities was 7.3 m (Fig. 7a2). These may be caused by the difference in estimation uncertainty between two products in the world. By comparing with reference data and VHR images, the GHSL-H product appears to underestimate building heights of large cities (Figs. 7b1&b3) and depict the hysteretic situation of building height in newly developed areas (Figs. 7c1&c4). Our product can provide a more detailed distribution of urban building heights. However, our product may underestimate building heights in low-latitude regions, probably due to shadows shortening with the latitude decrease (Figs. 7c3&c6),

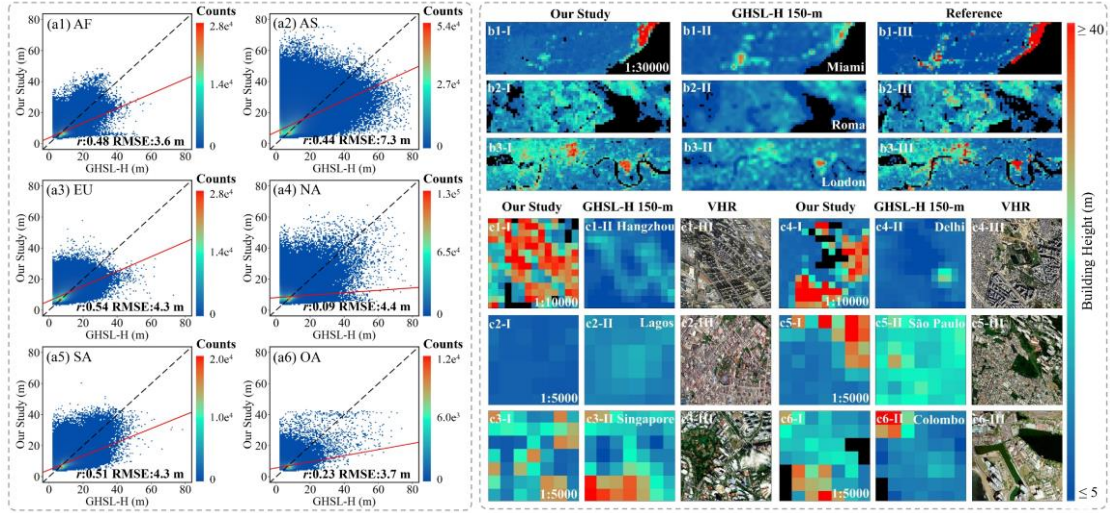

**Fig. 7.** Comparisons of the GHSL-H product and our product with a 150-m grid size at the pixel level for six continents (a1-a6); local views of two products and the reference building height data (b1-b3); close-up views of two products with the comparison of VHR images (c1-c6).

For further comparison, we downscaled our map and the GHSL-H product to the 1-km resolution in line with the Li et al. (2022) product for cities greater than 100 km$^2$ using the GUB2018. The three products showed little differences in building height distribution for typical developed cities (Figs. 8b3, b4, b6). However, there are significant differences between the three products when it comes to Africa and Asia (Figs. 8a1&a2). Our preliminary comparison suggests that there seems certain underestimation of the GHSL-H product and Li et al. (2022) product. For instance, quick scans show that Li et al. (2022) appeared to deviate from the building height distribution interpreted from the VHR images in Cairo (Fig. 8b1); the GHSL-H product and Li et al. (2022) product may not be able to capture the rapid urbanization in central Chengdu (Fig. 8b2); and the GHSL-H product seemed to overestimate building height in the mountainous area of Rio (Figs. 8b5).

On the whole, there was a small proportion of low values in our estimated results and a similar frequency distribution of three products in the range of mid-to-high values at each continent (Figs. 8a1-a6). Systematic studies are needed to assess the performance of the existing urban building height with details.

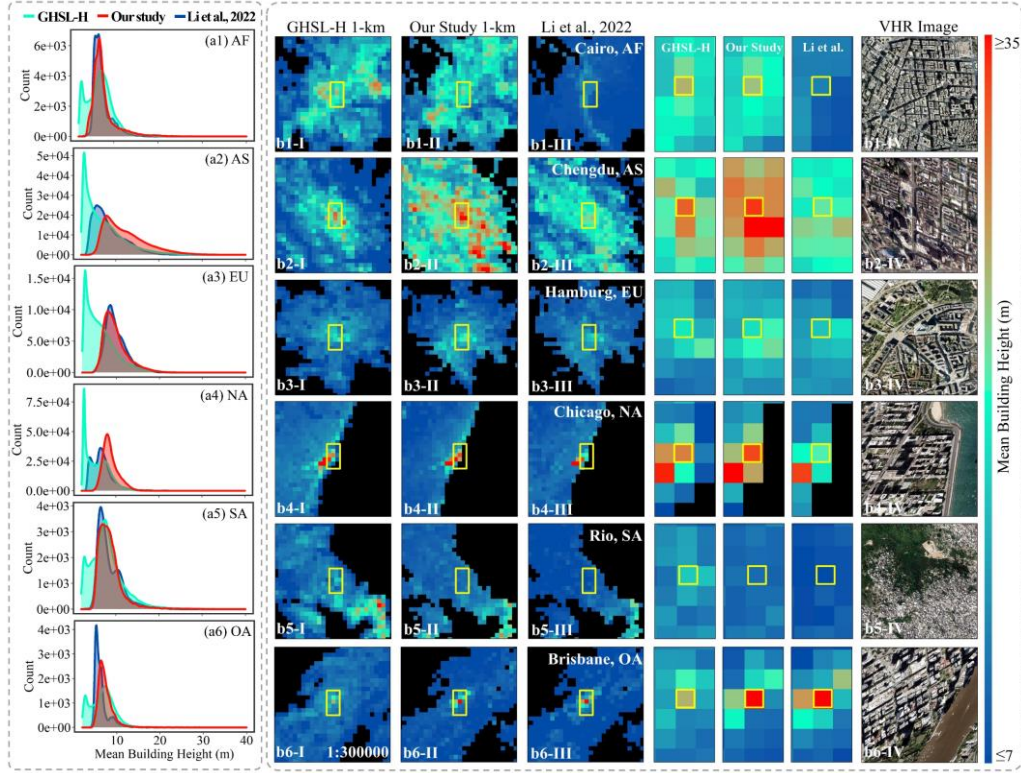

**Fig. 8.** Histograms of three urban building height datasets including GHSL-H, Li et al. (2022), and our products at the 1-km grid size (a1-a6); comparisons of three products for six cities (b1-b6) around the world with the Bing Image Map (b1-b6-III).

## 4. Conclusion

This study demonstrated the mapping processes of urban building heights at the global scale based on the integration of the GEDI-based building height samples and the RS-derived feature using the random forest models for 68 subregions. We further explored the potential of the GEDI data of retrieving building height in retrieving building heights and making building height samples around the world. The GEDI-derived urban building height map with a 150-m scale around the year 2020 included about 65,450 urban areas and performed good agreement with the reference building height data (i.e., Pearson's $r$ = 0.71, RMSE = 4.60 m). Besides these, this product showed spatial details of urban building height distribution with less blur and accurate status of newly developed areas. Using building height samples generated from spaceborne lidar makes updating this product more convenient in the next work. We expect this global building height product to boost future urban studies across many fields including climate, environmental, ecological, and social sciences.

# References:

Abrams, M., Crippen, R., & Fujisada, H. (2020). ASTER global digital elevation model (GDEM) and ASTER global water body dataset (ASTWBD). *Remote Sensing, 12*, 1156

Abrams, M., Yamaguchi, Y., & Crippen, R. (2022). Aster Global dem (gdem) Version 3. *The International Archives of the Photogrammetry, Remote Sensing and Spatial Information Sciences, 43*, 593-598

Amani, M., Ghorbanian, A., Ahmadi, S.A., Kakooei, M., Moghimi, A., Mirmazloumi, S.M., Moghaddam, S.H.A., Mahdavi, S., Ghahremanloo, M., & Parsian, S. (2020). Google earth engine cloud computing platform for remote sensing big data applications: A comprehensive review. *IEEE Journal of Selected Topics in Applied Earth Observations and Remote Sensing, 13*, 5326-5350

Beck, H.E., Zimmermann, N.E., McVicar, T.R., Vergopolan, N., Berg, A., & Wood, E.F. (2018). Present and future Köppen-Geiger climate classification maps at 1-km resolution. *Scientific Data, 5*, 180214

Breiman, L., Cutler, A., Liaw, A., & Wiener, M. (2018). Package 'randomForest': Breiman and Cutler's Random Forests for Classification and Regression. https://cran.r-project.org/web/packages/randomForest

Cai, B., Shao, Z., Huang, X., Zhou, X., & Fang, S. (2023). Deep learning-based building height mapping using Sentinel-1 and Sentinel-2 data. *International Journal of Applied Earth Observation and Geoinformation, 122*, 103399

Cao, Y.X., & Huang, X. (2021). A deep learning method for building height estimation using high-resolution multi-view imagery over urban areas: A case study of 42 Chinese cities. *Remote Sensing of Environment, 264*, 112590-112609

Chen, P., Huang, H., Liu, J., Wang, J., Liu, C., Zhang, N., Su, M., & Zhang, D. (2023). Leveraging Chinese GaoFen-7 imagery for high-resolution building height estimation in multiple cities. *Remote Sensing of Environment, 298*, 113802

Copernicus Global Land Service (2013). This publication has been prepared using European Union's Copernicus Land Monitoring Service information. https://doi.org/10.2909/42690e05-edf4-43fc-8020-33e130f62023, https://land.copernicus.eu/en/products/urban-atlas/building-height-2012

Dandabathula, G., Sitiraju, S.R., & Jha, C.S. (2021). Retrieval of building heights from ICESat-2 photon data and evaluation with field measurements. *Environmental Research: Infrastructure and Sustainability, 1*, 011003

Dimoudi, A., & Nikolopoulou, M. (2003). Vegetation in the urban environment: microclimatic analysis and benefits. *Energy and buildings, 35*, 69-76

Dubayah, R., Armston, J., Healey, S.P., Bruening, J.M., Patterson, P.L., Kellner, J.R., Duncanson, L., Saarela, S., Ståhl, G., & Yang, Z. (2022). GEDI launches a new era of biomass inference from space. *Environmental Research Letters, 17*, 095001

Dubayah, R., Blair, J.B., Goetz, S., Fatoyinbo, L., Hansen, M., Healey, S., Hofton, M., Hurtt, G., Kellner, J., & Luthcke, S. (2020). The Global Ecosystem Dynamics Investigation: High-resolution laser ranging of the Earth's forests and topography. *Science of remote sensing, 1*, 100002

Dubayah, R., Hofton, M., Blair, J., Armston, J., Tang, H., & Luthcke, S. (2021). GEDI L2A Elevation and Height Metrics Data Global Footprint Level V002 [Data set]. In, *NASA EOSDIS Land Processes Distributed Active Archive Center*

Esch, T., Brzoska, E., Dech, S., Leutner, B., Palacios-Lopez, D., Metz-Marconcini, A., Marconcini, M., Roth, A., & Zeidler, J. (2022). World Settlement Footprint 3D-A first three-dimensional survey of the global building stock. *Remote Sensing of Environment, 270*, 112877

Frantz, D., Schug, F., Okujeni, A., Navacchi, C., Wagner, W., van der Linden, S., & Hostert, P. (2021). National-scale mapping of building height using Sentinel-1 and Sentinel-2 time series. *Remote Sensing of Environment, 252*, 112128

Geiß, C., Schrade, H., Pelizari, P.A., & Taubenböck, H. (2020). Multistrategy ensemble regression for mapping of built-up density and height with Sentinel-2 data. *ISPRS Journal of Photogrammetry and Remote Sensing, 170*, 57-71

Gorelick, N., Hancher, M., Dixon, M., Ilyushchenko, S., Thau, D., & Moore, R. (2017). Google Earth Engine: Planetary-scale geospatial analysis for everyone. *Remote Sensing of Environment, 202*, 18-27

Haralick, R.M., Shanmugam, K., & Dinstein, I.H. (1973). Textural features for image classification. *IEEE Transactions on systems, man, and cybernetics*, 610-621

He, T., Wang, K., Xiao, W., Xu, S., Li, M., Yang, R., & Yue, W. (2023). Global 30 meters spatiotemporal 3D urban expansion dataset from 1990 to 2010. *Scientific Data, 10*, 321

Herfort, B., Lautenbach, S., Porto de Albuquerque, J., Anderson, J., & Zipf, A. (2023). A spatio-temporal analysis investigating completeness and inequalities of global urban building data in OpenStreetMap. *Nature Communications, 14*, 3985

Kellner, J.R., Armston, J., & Duncanson, L. (2023). Algorithm theoretical basis document for GEDI footprint aboveground biomass density. *Earth and Space Science, 10*, e2022EA002516

Lao, J., Wang, C., Zhu, X., Xi, X., Nie, S., Wang, J., Cheng, F., & Zhou, G. (2021). Retrieving building height in urban areas using ICESat-2 photon-counting LiDAR data. *International Journal of Applied Earth Observation and Geoinformation, 104*, 102596-102606

Li, M., Wang, Y., Rosier, J.F., Verburg, P.H., & van Vliet, J. (2022). Global maps of 3D built-up patterns for urban morphological analysis. *International Journal of Applied Earth Observation and Geoinformation, 114*, 103048

Li, X., Gong, P., Zhou, Y., Wang, J., Bai, Y., Chen, B., Hu, T., Xiao, Y., Xu, B., & Yang, J. (2020a). Mapping global urban boundaries from the global artificial impervious area (GAIA) data. *Environmental Research Letters, 15*, 094044

Li, X., Zhou, Y., Gong, P., Seto, K.C., & Clinton, N. (2020b). Developing a method to estimate building height from Sentinel-1 data. *Remote Sensing of Environment, 240*, 111705

Ma, X., Zheng, G., Chi, X., Yang, L., Geng, Q., Li, J., & Qiao, Y. (2023). Mapping fine-scale building heights in urban agglomeration with spaceborne lidar. *Remote Sensing of Environment, 285*, 113392

Marconcini, M. (2021). The View from Space-How Cities are Growing. In: German Remote Sensing Data Center

Patino, J.E., & Duque, J.C. (2013). A review of regional science applications of satellite remote sensing in urban settings. *Computers, Environment and Urban Systems, 37*, 1-17

Pedersen Zari, M., MacKinnon, M., Varshney, K., & Bakshi, N. (2022). Regenerative living cities and the urban climate–biodiversity–wellbeing nexus. *Nature Climate Change, 12*, 601-604

Perini, K., & Magliocco, A. (2014). Effects of vegetation, urban density, building height, and atmospheric conditions on local temperatures and thermal comfort. *Urban Forestry & Urban Greening, 13*, 495-506

Pesaresi, M., & Politis, P. (2023). GHS-BUILT-H R2023A - GHS building height, derived from AW3D30, SRTM30, and Sentinel2 composite (2018). European Commission, Joint Research Centre (JRC) [Dataset]. doi: 10.2905/85005901-3A49-48DD-9D19-6261354F56FE

Potapov, P., Li, X., Hernandez-Serna, A., Tyukavina, A., Hansen, M.C., Kommareddy, A., Pickens, A., Turubanova, S., Tang, H., & Silva, C.E. (2021). Mapping global forest canopy height through integration of GEDI and Landsat data. *Remote Sensing of Environment, 253*, 112165

Rouse, J. (1973). Monitoring the vernal advancement and retrogradation of natural vegetation. *NASA/GSFCT Type II Report*

United Nations. (2022). World-Population-Prospects-2022. https://www.un.org/development/desa/pd/content/World-Population-Prospects-2022

Wang, Y., Li, X., Yin, P., Yu, G., Cao, W., Liu, J., Pei, L., Hu, T., Zhou, Y., & Liu, X. (2023). Characterizing annual dynamics of urban form at the horizontal and vertical dimensions using long-term Landsat time series data. *ISPRS Journal of Photogrammetry and Remote Sensing, 203*, 199-210

Wu, W.-B., Ma, J., Banzhaf, E., Meadows, M.E., Yu, Z.-W., Guo, F.-X., Sengupta, D., Cai, X.-X., & Zhao, B. (2023). A first Chinese building height estimate at 10 m resolution (CNBH-10 m) using multi-source earth observations and machine learning. *Remote Sensing of Environment, 291*, 113578

Xu, H. (2006). Modification of normalised difference water index (NDWI) to enhance open water features in remotely sensed imagery. *International journal of remote sensing, 27*, 3025-3033

Zha, Y., Gao, J., & Ni, S. (2003). Use of normalized difference built-up index in automatically mapping urban areas from TM imagery. *International journal of remote sensing, 24*, 583-594

Zhang, Z., Qian, Z., Zhong, T., Chen, M., Zhang, K., Yang, Y., Zhu, R., Zhang, F., Zhang, H., Zhou, F., Yu, J., Zhang, B., Lü, G., & Yan, J. (2022). Vectorized rooftop area data for 90 cities in China. *Scientific Data, 9*, 66

Zhao, Y., Wu, B., Li, Q., Yang, L., Fan, H., Wu, J., & Yu, B. (2023). Combining ICESat-2 photons and Google Earth Satellite images for building height extraction. *International Journal of Applied Earth Observation and Geoinformation, 117*, 103213

Zhou, Y., Li, X., Chen, W., Meng, L., Wu, Q., Gong, P., & Seto, K.C. (2022). Satellite mapping of urban built-up heights reveals extreme infrastructure gaps and inequalities in the Global South. *Proceedings of the National Academy of Sciences, 119*, e2214813119